\documentclass[conference]{IEEEtran}
\IEEEoverridecommandlockouts
\usepackage{cite}
\usepackage{amsmath,amssymb,amsfonts}
\usepackage{algorithmic}
\usepackage{graphicx}
\usepackage{textcomp}
\usepackage{xcolor}
\def\BibTeX{{\rm B\kern-.05em{\sc i\kern-.025em b}\kern-.08em
    T\kern-.1667em\lower.7ex\hbox{E}\kern-.125emX}}

\usepackage[caption=false]{subfig} 
\usepackage[inline]{enumitem} 
\usepackage{hyperref}
\usepackage{cleveref}
\usepackage{xurl}
\usepackage{array}
\usepackage{bm}

\usepackage{booktabs}
\usepackage{multirow}

\graphicspath{{figures/}} 

\begin{document}

\title{On the Joint Minimization of Regularization Loss Functions in Deep Variational Bayesian Methods for Attribute-Controlled Symbolic Music Generation}

\author{
\IEEEauthorblockN{Matteo Petten{\`o}}
\IEEEauthorblockA{
\textit{Politecnico di Milano}\\
Milan, Italy \\
matteo.petteno@mail.polimi.it}
\and
\IEEEauthorblockN{Alessandro Ilic Mezza}
\IEEEauthorblockA{
\textit{Politecnico di Milano}\\
Milan, Italy \\
alessandroilic.mezza@polimi.it}
\and
\IEEEauthorblockN{Alberto Bernardini}
\IEEEauthorblockA{
\textit{Politecnico di Milano}\\
Milan, Italy \\
alberto.bernardini@polimi.it}
}

\maketitle
  
\begin{abstract}
Explicit latent variable models provide a flexible yet powerful framework for data synthesis, enabling controlled manipulation of generative factors. With latent variables drawn from a tractable probability density function that can be further constrained, these models enable continuous and semantically rich exploration of the output space by navigating their latent spaces. Structured latent representations are typically obtained through the joint minimization of regularization loss functions. In variational information bottleneck models, reconstruction loss and Kullback-Leibler Divergence (KLD) are often linearly combined with an auxiliary Attribute-Regularization (AR) loss. However, balancing KLD and AR turns out to be a very delicate matter. When KLD dominates over AR, generative models tend to lack controllability; when AR dominates over KLD, the stochastic encoder is encouraged to violate the standard normal prior. We explore this trade-off in the context of symbolic music generation with explicit control over continuous musical attributes. We show that existing approaches struggle to jointly minimize both regularization objectives, whereas suitable attribute transformations can help achieve both controllability and regularization of the target latent dimensions. 
\end{abstract}


\section{Introduction}
\label{sec:introduction}
Symbolic music generation has witnessed remarkable advancements in recent years with the rise of deep latent variable models~\cite{robertsHierarchicalLatentVector2018, gillick2019groove, engel2018latentconstraints,mittal2021symbolicdiffusion, midime}, yet the challenge of controlling high-level musical attributes at inference time remains an active area of research. 
Latent variable models can be broadly categorized into \textit{explicit} density models~\cite{wuBridgingExplicitImplicit2021}, such as Variational Autoencoders (VAEs), Variational Information Bottleneck (VIB) models, normalizing flows, and diffusion models, and \textit{implicit} density models, notably Generative Adversarial Networks (GANs).

Explicit density models offer the advantage of tractable likelihood estimation and a well-structured latent space. 
In particular, deep variational Bayesian methods~\cite{kingmaAutoEncodingVariationalBayes2014} learn smooth and continuous representations of musical data~\cite{robertsHierarchicalLatentVector2018, gillick2019groove}, leading to meaningful interpolation and targeted manipulation of specific characteristics of the output~\cite{Roberts2018LearningLR}. 
Moreover, VAEs~\cite{higgins2017beta} and VIB models~\cite{alemiDeepVariationalInformation2017} are able to learn disentangled representations, i.e., having independent and interpretable latent factors, which, in turn, allow individual musical attributes to be adjusted without unintentionally altering others.
This is often accomplished through supervised multi-task learning techniques~\cite{patiAttributebasedRegularizationLatent2021, mezzaLatentRhythmComplexity2023, hadjeres2017glsr_vae, attri-vae, banar2023tool}, which structure the latent space by ensuring that specific attributes are encoded so as to be proportional to changes along designated dimensions. These approaches entail introducing auxiliary attribute-regularization objectives alongside the reconstruction loss, ensuring data fidelity, and Kullback-Leibler Divergence (KLD), which constrains the latent space to be continuous and evenly dense, in addition to enabling efficient and reliable sampling from a predefined prior distribution.

To the best of our knowledge, however, the trade-off between these regularization losses has not been thoroughly explored in the literature, with training recipes being often provided without an accompanying analysis of the ensuing behavior of the regularized dimensions. In this work, by focusing on high-level musical attributes such as Contour, Pitch Range, and Rhythm Complexity, we show that existing attribute-controlled symbolic music generation models are highly sensitive to hyperparameter tuning and tend to fail in satisfying both objectives at the same time. Furthermore, we show that invertible attribute mappings based on power transforms may help mitigate the issue.

\section{Related Work}
\label{sec:background}

Let $X$, $Y$, $Z$ be random variables. The Information Bottleneck (IB)~\cite{tishby1999ib} optimization problem is given by~\cite{alemiDeepVariationalInformation2017} $\max_{\theta}\,I_{\theta}(Z, Y) - \beta I_{\theta}(Z, X)$, where $I(\cdot, \cdot)$ denotes the mutual information. The IB framework entails finding $p_{\theta}(z|x)$ parameterized by $\theta$ that defines a compressed representation of $X$, i.e., the latent variable $Z$, which retains as much information as possible about $Y$ while discarding irrelevant information about $X$. This trade-off is controlled by the Lagrange multiplier $\beta \geq 0$.

The VIB approach~\cite{alemiDeepVariationalInformation2017} 
finds a variational lower bound on the IB objective, resulting in the minimization of the following loss function:
\begin{equation}
\label{eq:vib_objective} 
    \mathcal{L}_{\text{VIB}} =  -\mathbb{E}_{p_{\theta}(z|x)}\left[\log q_{\phi}(y|z)\right] + \beta D_{\text{KL}} \left[ p_{\theta}(z|x) \parallel r(z) \right],
\end{equation}
where $p_\theta(z|x)$ is a stochastic encoder, $x$ is the encoder input, $q_\phi(y|z)$ is a variational approximation of $p(y|z)$ parameterized by the weights $\phi$ of a decoder network, $y$ is the decoder output, $r(z)$ is the variational approximation of the marginal distribution $p(z)$, and $D_{\text{KL}}[\cdot\Vert\cdot]$ denotes the KLD.
If $Y = X$, \eqref{eq:vib_objective} coincides with the classic $\beta$-VAE objective~\cite{higgins2017beta}.
In most practical cases, $p_{\theta}(z|x)$ and $r(z)$ are chosen to be multivariate normal distributions, namely, $p_{\theta}(z|x) \sim \mathcal{N}(\bm{\mu}_{z|x}, \bm{\Sigma}_{z|x})$ and $r(z) \sim \mathcal{N}(\mathbf{0}, \mathbf{I})$.

With little loss of generality, we will focus here on the one-dimensional continuous case where an \textit{attribute} is defined as a scalar $a\in\mathcal{A}$ that can be algorithmically computed from an instance $x$,  i.e., there exists $f: x \mapsto f(x)$ such that $f(x)=a$, with $\mathcal{A}\subseteq\mathbb{R}$ the image of $f$.

To develop \textit{attribute-controlled} generative models,~\cite{mezzaLatentRhythmComplexity2023, patiAttributebasedRegularizationLatent2021} propose to encode $a$ onto the $i$-th dimension of the latent space by adding a supervised Attribute Regularization (AR) term to \eqref{eq:vib_objective}, i.e.,
\begin{equation}
\label{eq:ar_vib_objective} 
    \mathcal{L}_\text{AR-VIB} =  \mathcal{L}_\text{VIB} + \gamma \mathcal{L}_\text{AR},
\end{equation}
where $\gamma \ge 0$ is a tunable hyperparameter.

In~\cite{mezzaLatentRhythmComplexity2023}, Mezza et al.\ define such a regularization term as
\begin{align}
\label{eq:beta_vae_default_attr_reg_term}
    \mathcal{L}_{\text{AR}} =  \operatorname{MAE}(z_i,\, \tilde{a}),
\end{align}
where $\operatorname{MAE}$ denotes the mean absolute error, and $\tilde{a}$ is the z-score of the attribute $a$.
This simple pointwise error approach proves effective as long as $a$ is normally distributed~\cite{mezzaLatentRhythmComplexity2023}. 
If not, the regularization losses in \eqref{eq:ar_vib_objective} turn out to have contrasting objectives. 
If $\mathcal{L}_{\text{AR}}$ dominates over the KLD, the learned posterior $p_\theta(z_i|x)$ tends to na{\"i}vely approximate the attribute's sample distribution $p(a)$ and disregard the target prior $r(z_i)$, with the covariance $\bm{\Sigma}_{z|x}$ collapsing toward a zero matrix. Conversely, if the KLD dominates over $\mathcal{L}_{\text{AR}}$, then $p_\theta(z_i|x)$ follows $r(z_i)$ at the cost of poor attribute regularization.

In~\cite{patiAttributebasedRegularizationLatent2021}, Pati and Lerch propose a novel term that relaxes the regularization constraint by enforcing a monotonic relationship between $a$ and $z_i$, i.e., 
\begin{align}
\label{eq:beta_vae_sign_attr_reg_term}
    \mathcal{L}_{\text{AR}} =  \operatorname{MAE} \left( \tanh(\delta\, \mathbf{D}_\mathrm{z}),\,  \operatorname{sign}(\mathbf{D}_\mathrm{a}) \right),
\end{align}
where $\mathbf{D}_\mathrm{z}$ and $\mathbf{D}_\mathrm{a}$ are the pairwise distance matrices between $z_i$ and $a$ of all samples in a given mini-batch, respectively, and $\delta>0$ is a hyperparameter controlling the spread of the posterior distribution. 
Since enforcing an arbitrary monotonic relationship may still result in a distribution diverging from the chosen prior,
though, using \eqref{eq:beta_vae_sign_attr_reg_term} requires careful balancing of $\beta$ and $\gamma$ in~\eqref{eq:ar_vib_objective} in order to learn $\theta$ such that $p_\theta(z_i|x)$ approximates $r(z_i)$. In fact, the posterior is encouraged to be Gaussian only as long as the KLD is minimized, which, in turn, may negatively affect attribute control.


\section{Power Transform-Based Regularization}
\label{sec:methodology}





In the previous section, we outlined practical problems arising from the minimization of~\eqref{eq:ar_vib_objective} when having two conflicting terms in $\mathcal{L}_\text{AR}$ and $D_{\text{KL}} \left[ p_{\theta}(z|x) \parallel r(z) \right]$. In fact, when the prior and the attribute distribution are too dissimilar, we will later show that $p_\theta(z_i|x)$ fails to jointly model $p(a)$ and $r(z_i)$.
However, we argue that, the opposite being true, a simple distance measure may prove sufficient as a regularizer. 
We can achieve this by means of a parametric isomorphism $T_{\lambda}: \mathcal{A}\rightarrow\mathbb{R}$ such that $p(T_{\lambda}(a))$ is (at least approximately) distributed like $r(z_i)$. Hence, we define the following regularization loss:
\begin{equation} 
\label{eq:ar_vib_nf_reg}
    \mathcal{L}_{\text{AR}} = \operatorname{MAE}(z_i,\, T_{\lambda}(a)).
\end{equation}

By requiring $T_{\lambda}$ to be invertible, we can seamlessly go back and forth between the latent space and the input space; this enables attribute manipulation within the original, human-interpretable domain, while ensuring the consistency of the component-wise marginals of the posterior.

In case $r(z_i)\sim\mathcal{N}(0,1)$, $T_\lambda$ has to transform data so as to make them more normal-like. We refer to this process as \textit{data Gaussianization}. Gaussianization can be implemented using both parametric~\cite{boxAnalysisTransformations1964, yeoNewFamilyPower2000} and non-parametric methods~\cite{chenGaussianization2000, laparraIterativeGaussianizationICA2011}. In the one-dimensional case, a simple approach is to use a family of functions that define monotonic data transformations, known in statistics as Power Transforms (PT). 

First introduced in~\cite{boxAnalysisTransformations1964}, one of the most widely used PT is the Box-Cox transformation, whose two-parameter formulation is
\begin{equation}
\label{eq:two_param_box_cox}
    g_{\lambda_1, \lambda_2}(u) = 
    \begin{cases}
        \frac{(u + \lambda_2)^{\lambda_1} -1 }{\lambda_1} & \text{if}\ \lambda_1 \neq 0, \\
        \ln(u + \lambda_2) & \text{if}\ \lambda_1 = 0,
    \end{cases}
\end{equation}
where the shift parameter $\lambda_2\ge0$ extends the domain of $g$ from $\left\{u\in\mathbb{R}\,\vert\,u>0\right\}$ to $\left\{u\in\mathbb{R}\,\vert\,u>-\lambda_2\right\}$.\footnote{In this study, we only consider attributes that, by construction, are bounded from below. If not, the Box-Cox transformation could be replaced, e.g., by the Yeo-Johnson transformation~\cite{yeoNewFamilyPower2000}, which is defined on all of~$\mathbb{R}$.}

The proposed transformation of the attribute distribution applies~\eqref{eq:two_param_box_cox} followed by a Batch Normalization layer with scale and shift parameters set to one and zero, respectively, i.e., 
\begin{equation}
\label{eq:batch_norm}
    \operatorname{BN}_{1, 0}(u) = \frac{u - \mu}{\sqrt{\sigma^2 + \epsilon}},
\end{equation}
where $\epsilon$ is a small constant, and $\mu$ and $\sigma$ correspond to the mean and standard deviation of the current mini-batch during training, and their moving average estimates across all batches at inference time. 
Thus, the whole transformation is given by the composition
\begin{equation} 
\label{eq:pt_gaussianization}
    T_{\lambda}(a) = \left( \operatorname{BN}_{1, 0}\, \circ\ g_{\lambda_1, \lambda_2}\right)(a).
\end{equation}

The transformation parameters $\lambda=\{\lambda_1, \lambda_2\}$ are determined using training data prior to the training phase, rather than being learned alongside the VIB parameters. Specifically, the shift parameter $\lambda_2$ is found through a grid search, and for each value in the grid, the power parameter $\lambda_1$ is estimated by minimizing the negative log-likelihood using Brent's method~\cite{brentAlgorithmGuaranteedConvergence1971}.
The optimal pair $(\lambda_1$, $\lambda_2)$ is selected by picking the parameters for which the transformed distribution has the lowest negentropy, which can be approximated by 
$J(\tilde{v}) \approx \left(\mathbb{E}\left[\psi(\tilde{v})\right] - \mathbb{E}\left[\psi(\nu)\right]\right)^2$~\cite{hyvarinenIndependentComponentAnalysis2000}, where $\tilde{v}$ is a standardized instance of a random variable~$V$, $\nu \sim \mathcal{N}(0, 1)$, and $\psi$ is a nonquadratic function that we chose to be $\psi(u) = -\exp(-u^2/\,2)$~\cite{hyvarinenIndependentComponentAnalysis2000}.
Once computed, $\{\lambda_1$, $\lambda_2\}$ is kept fixed during both training and inference, so applying \eqref{eq:pt_gaussianization} incurs no additional cost.

Looking back at \eqref{eq:pt_gaussianization}, we may notice that $T_\lambda$ is the composition of differentiable, monotone, and invertible functions. As such, we may interpret $T_\lambda^{-1}$ as a special case of one-dimensional normalizing flows where a nonnormal distribution $p(a)$ is obtained by letting the simple distribution $r(z_i)$ flow through the diffeomorphism. While we do not explicitly require $T_\lambda(a)$ to be differentiable, using diffeomorphic attribute transformations would ultimately allow to learn $\lambda$ via stochastic gradient descent. We leave this study for future work.

\section{Evaluation}
\label{sec:evaluation}

\begin{figure*}[ht]
    \centering
    \subfloat[NM \label{fig:contour_plots_gamma=0.001_NM}]{
        \includegraphics[width=0.22\linewidth]{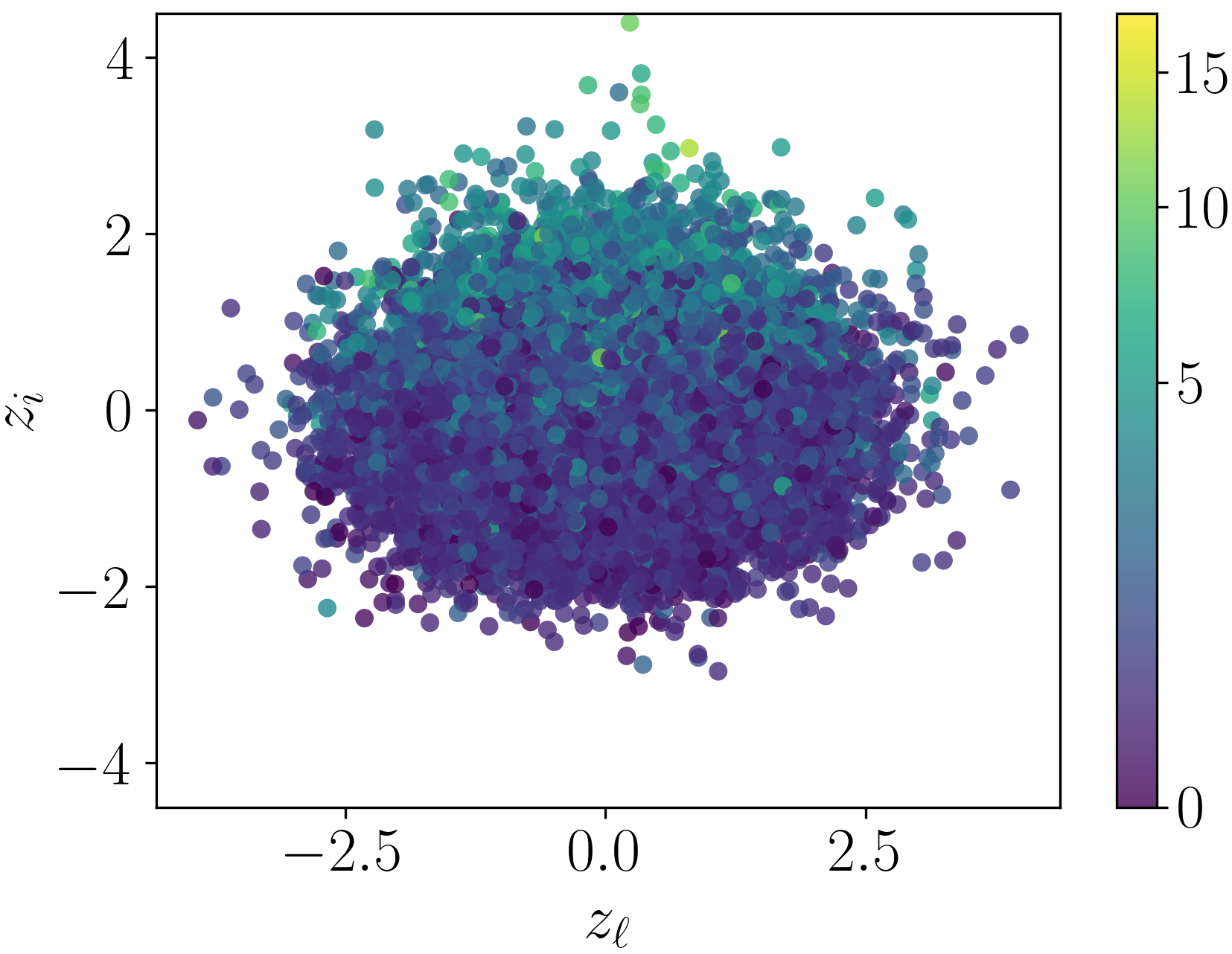}
    }\hspace{1cm}
    \subfloat[P\&L \label{fig:contour_plots_gamma=0.001_PL}]{
		\includegraphics[width=0.22\linewidth]{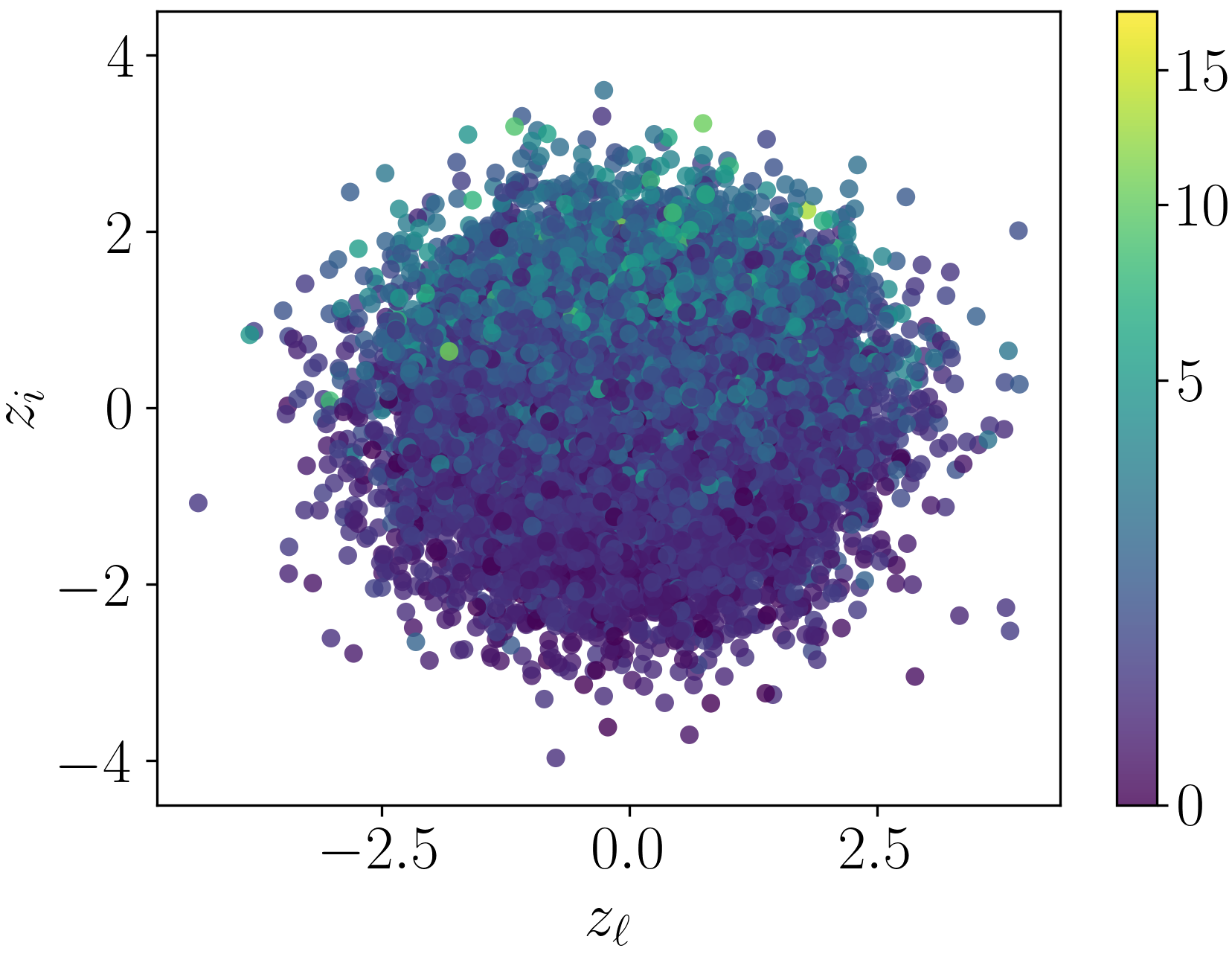}
    }\hspace{1cm}
    \subfloat[PT \label{fig:contour_plots_gamma=0.001_PT}]{
		\includegraphics[width=0.22\linewidth]{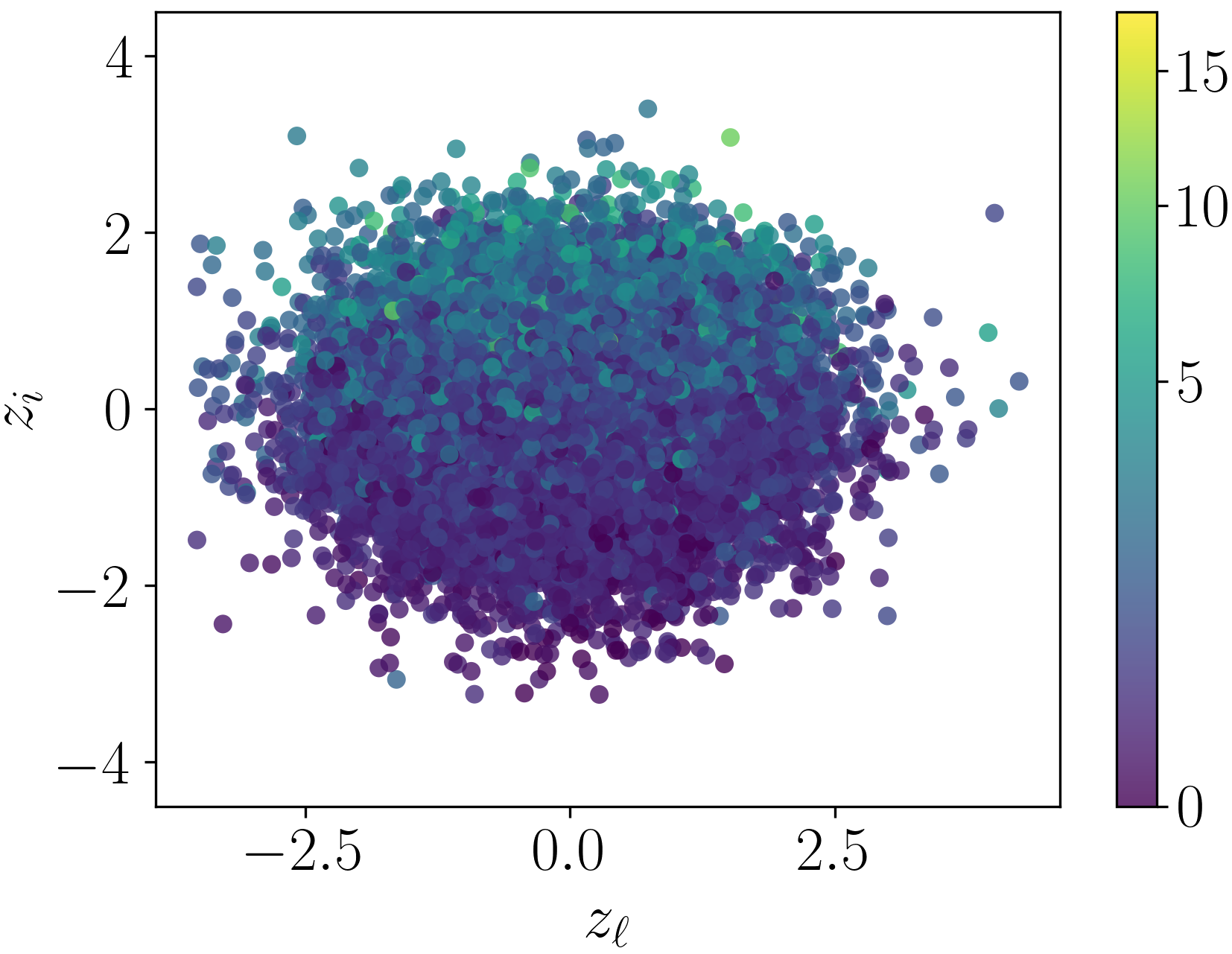}
    }\\[-0.5em]
    \subfloat[NM \label{fig:contour_plots_gamma=0.001_NM_OA}]{
		\includegraphics[width=0.22\linewidth]{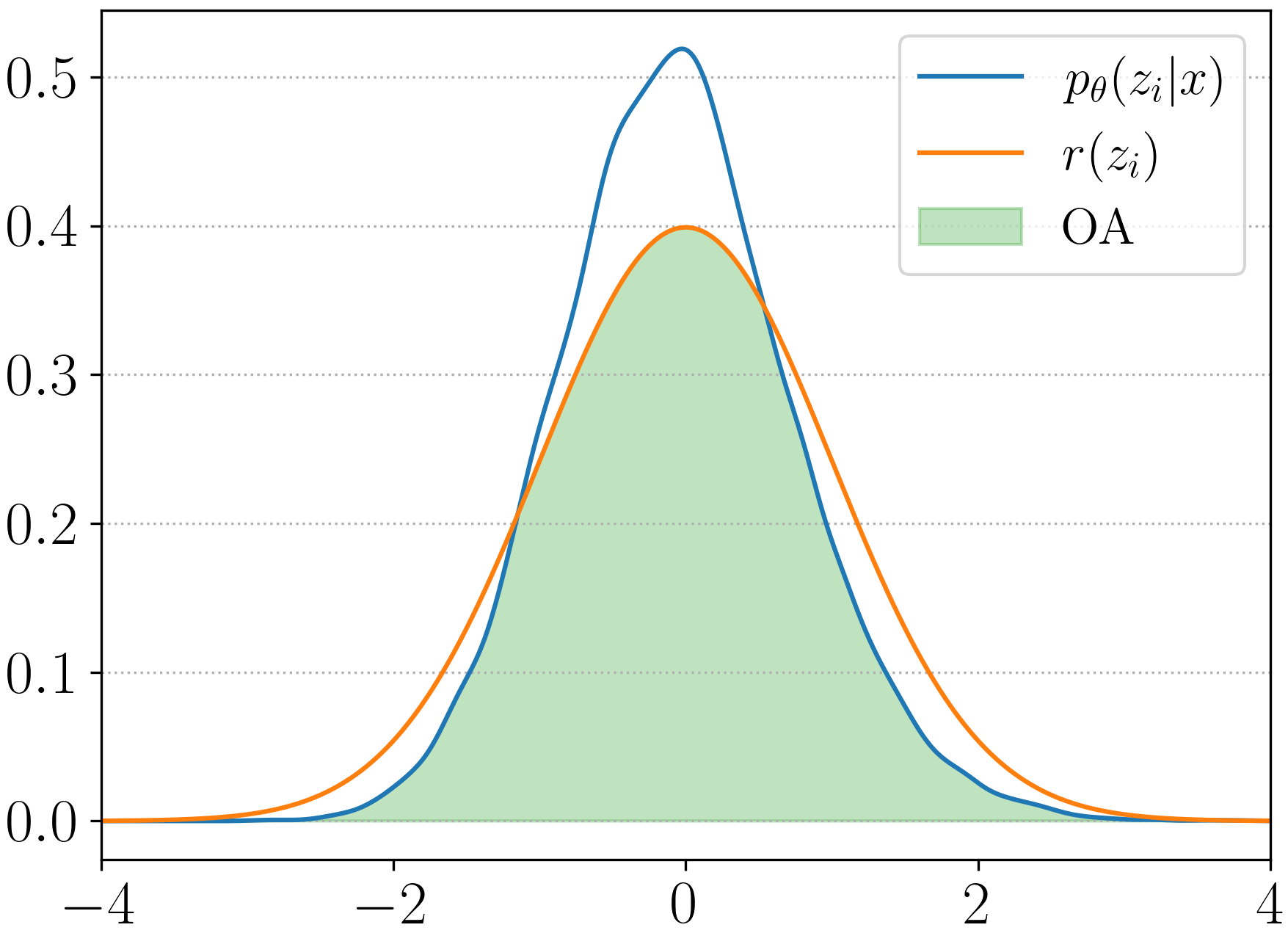}
    }\hspace{1cm}
    \subfloat[P\&L \label{fig:contour_plots_gamma=0.001_PL_OA}]{
		\includegraphics[width=0.22\linewidth]{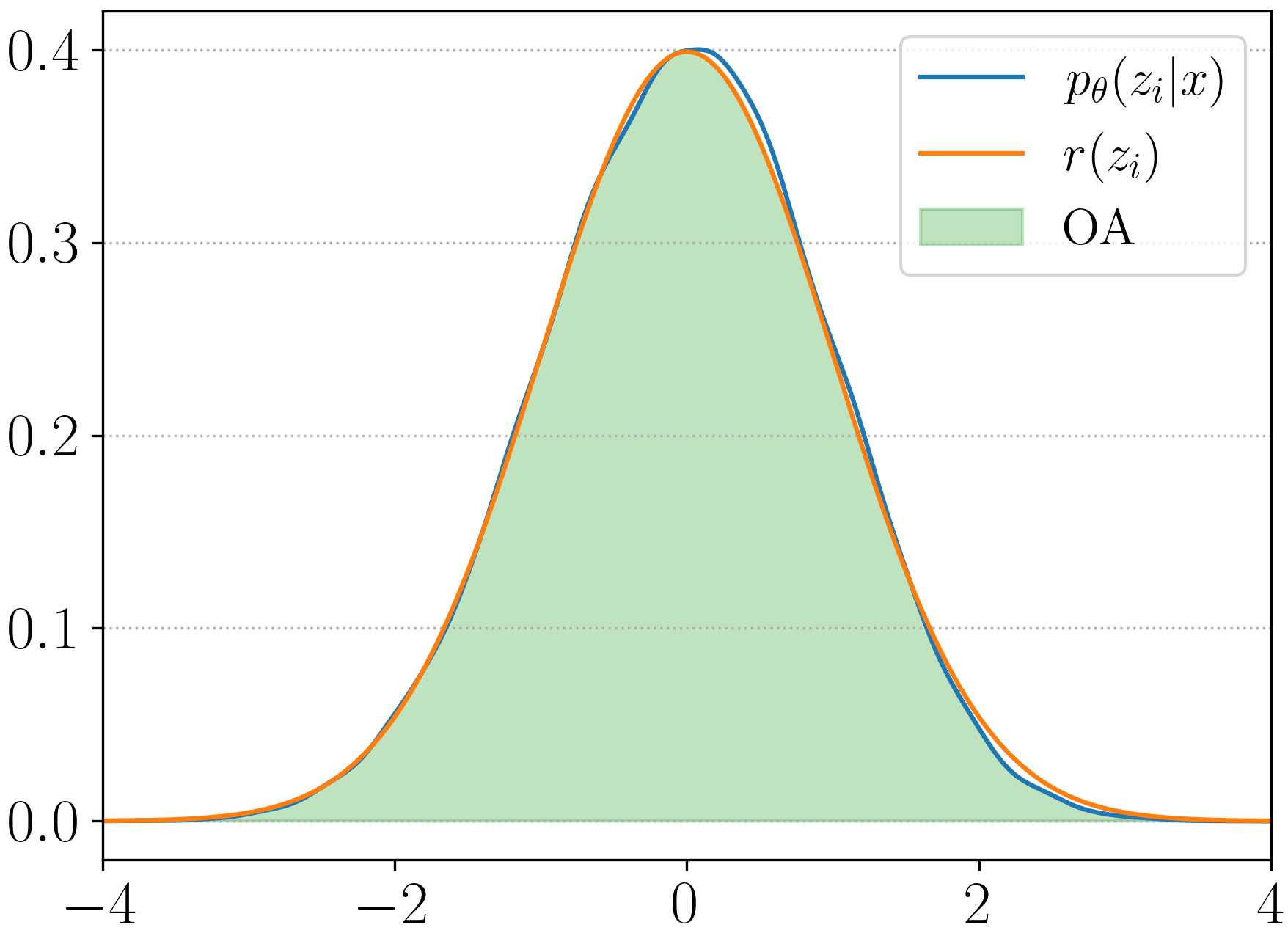}
    }\hspace{1cm}
    \subfloat[PT \label{fig:contour_plots_gamma=0.001_PT_OA}]{
		\includegraphics[width=0.22\linewidth]{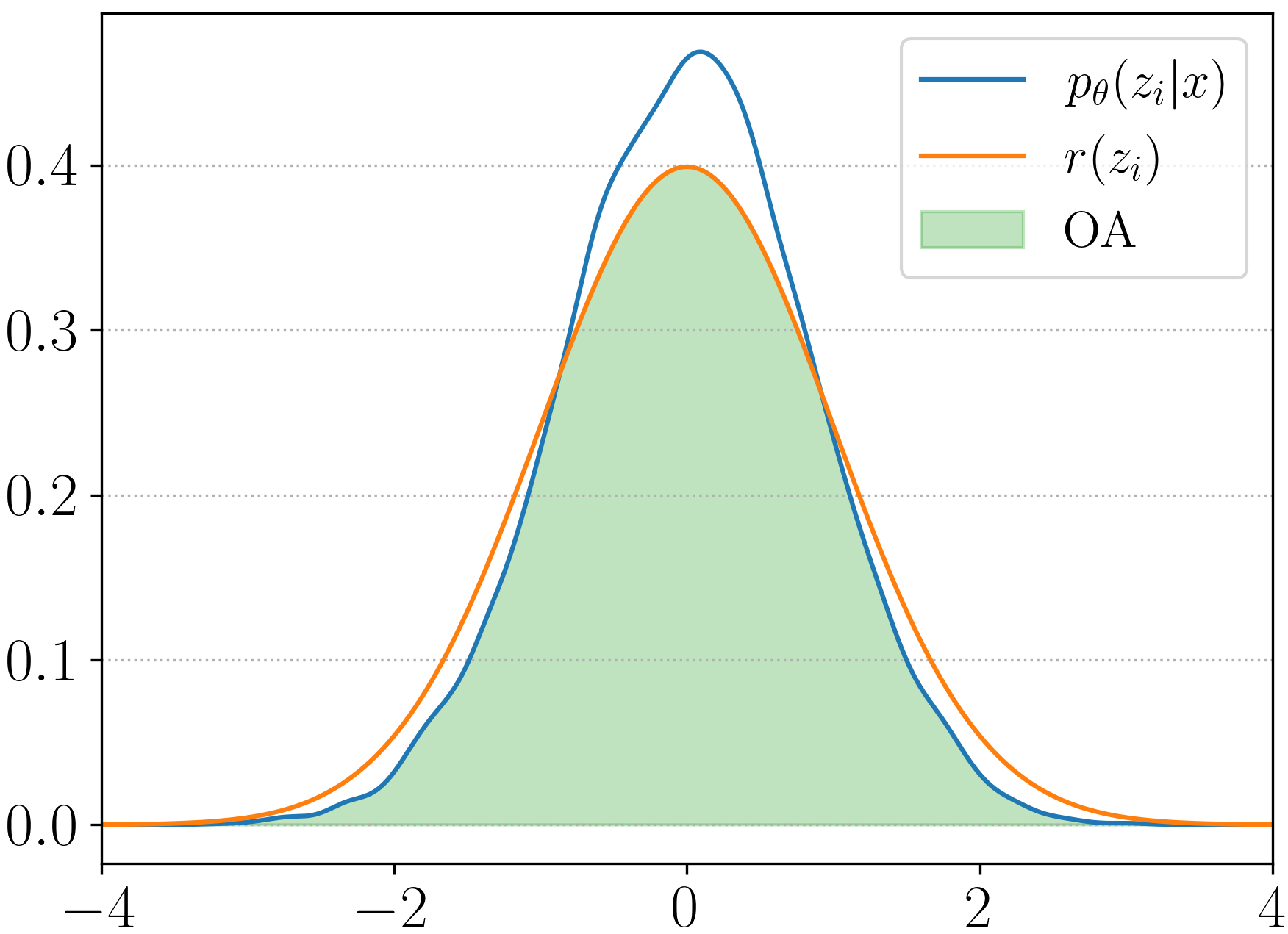}
    }\\
\caption{\textbf{Contour} attribute regularization with $\beta\rightarrow10^{-3}$ and $\gamma=10^{-3}$}
\label{fig:contour_plots_gamma=0.001}
\vspace{0.25em}
        \subfloat[NM \label{fig:contour_plots_gamma=1.0_NM}]{
		\includegraphics[width=0.22\linewidth]{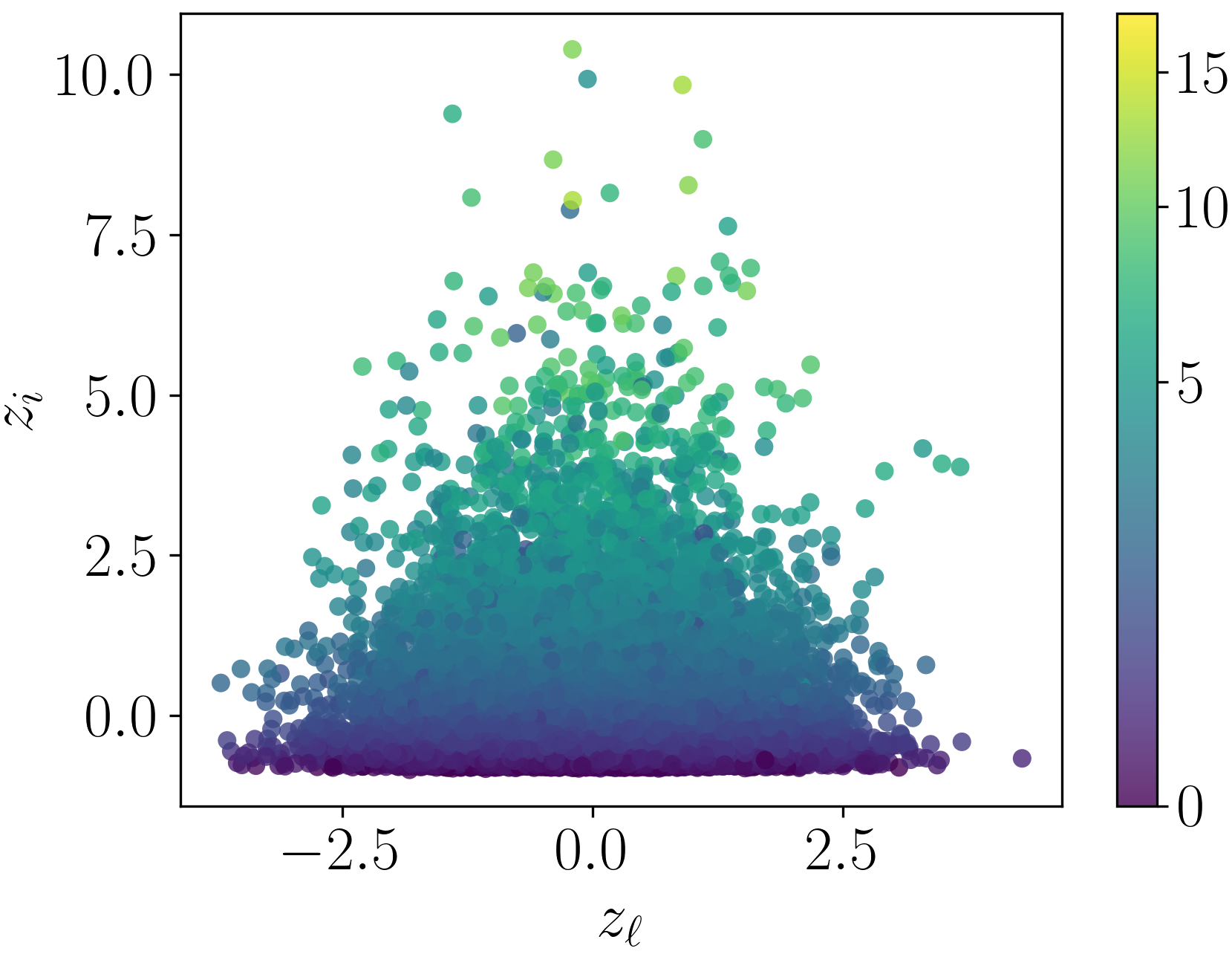}
        }\hspace{1cm}
        \subfloat[P\&L \label{fig:contour_plots_gamma=1.0_PL}]{
		\includegraphics[width=0.22\linewidth]{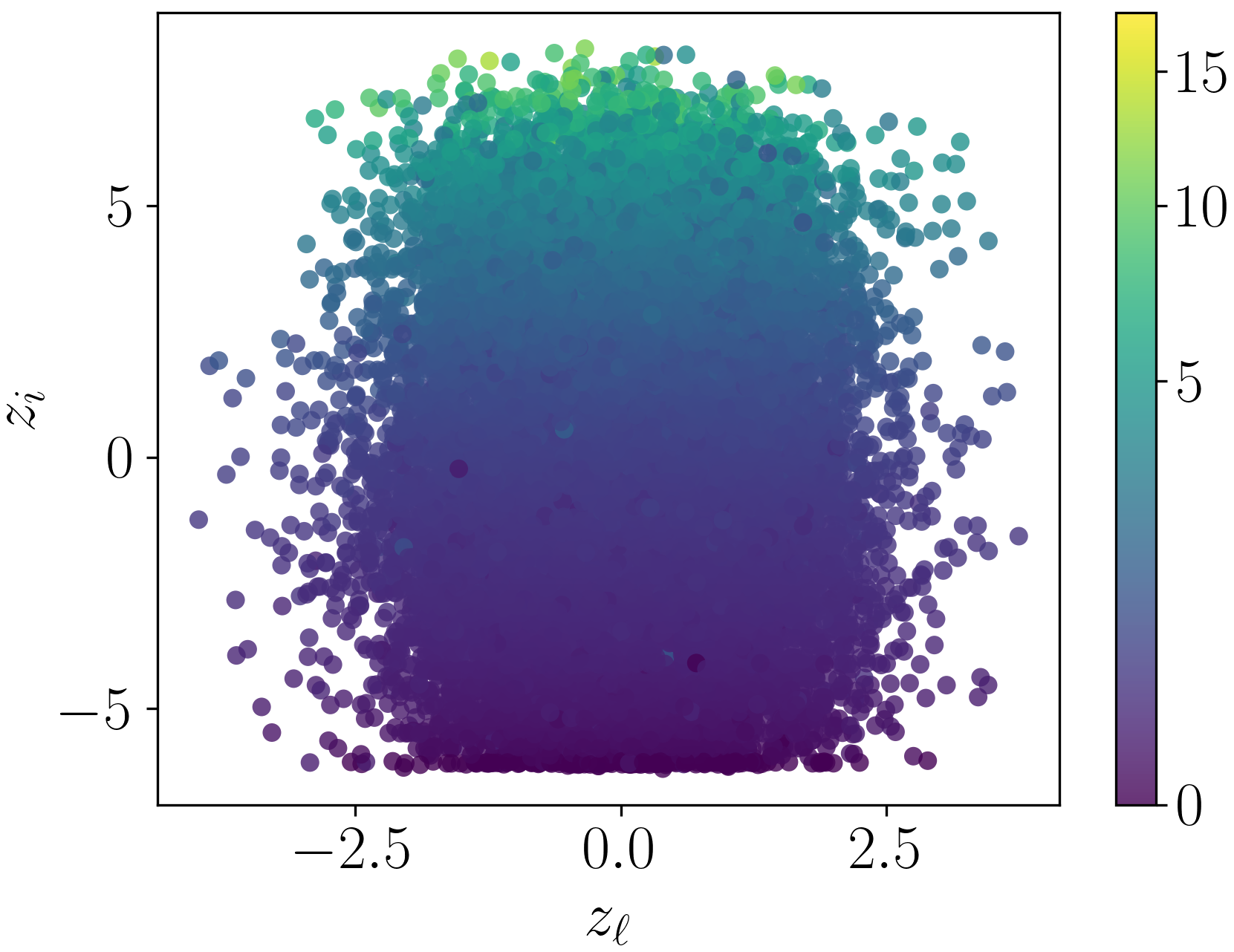}
        }\hspace{1cm}
        \subfloat[PT \label{fig:contour_plots_gamma=1.0_PT}]{
		\includegraphics[width=0.22\linewidth]{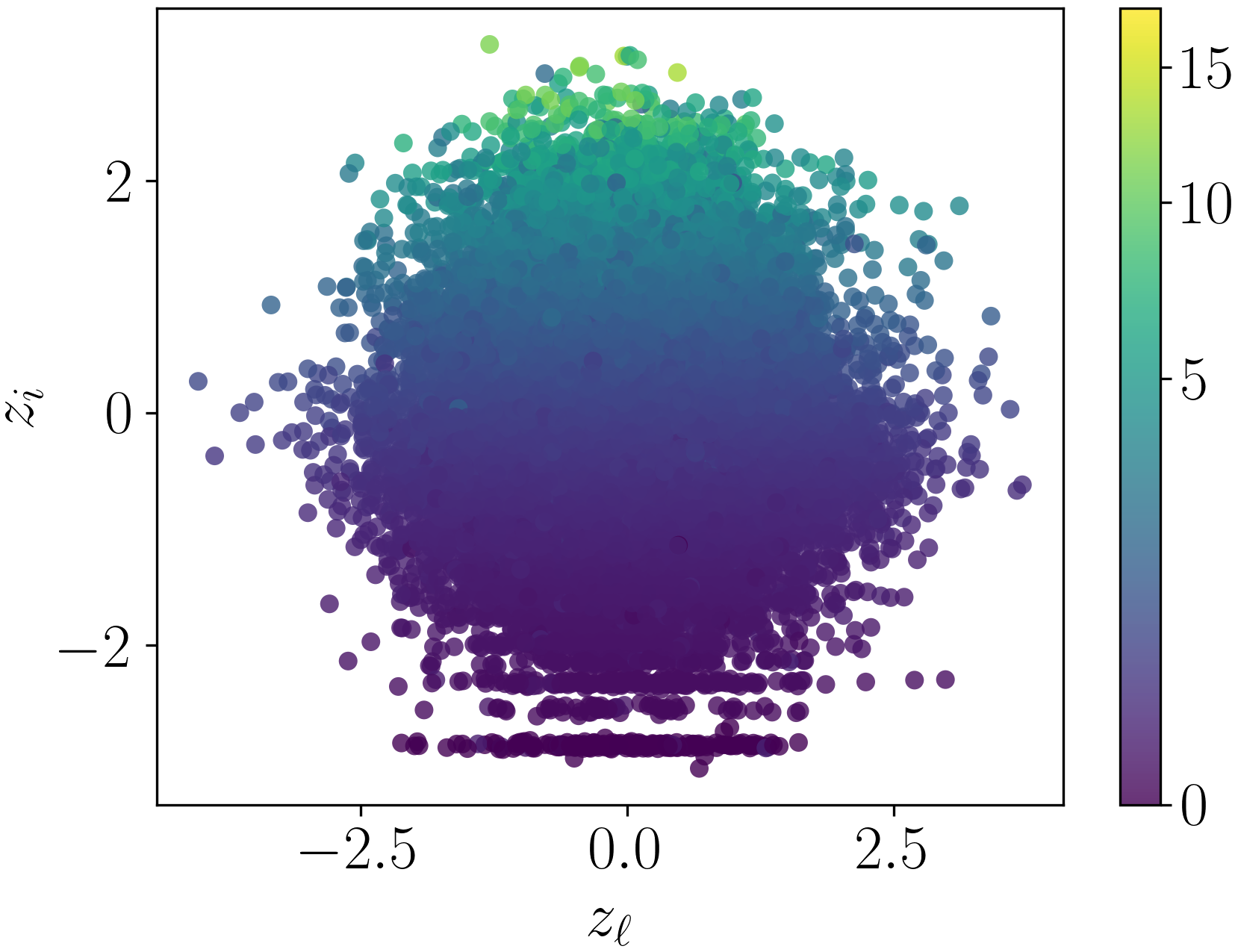}
        }\\[-0.5em]
        \subfloat[NM \label{fig:contour_plots_gamma=1.0_NM_OA}]{
		\includegraphics[width=0.22\linewidth]{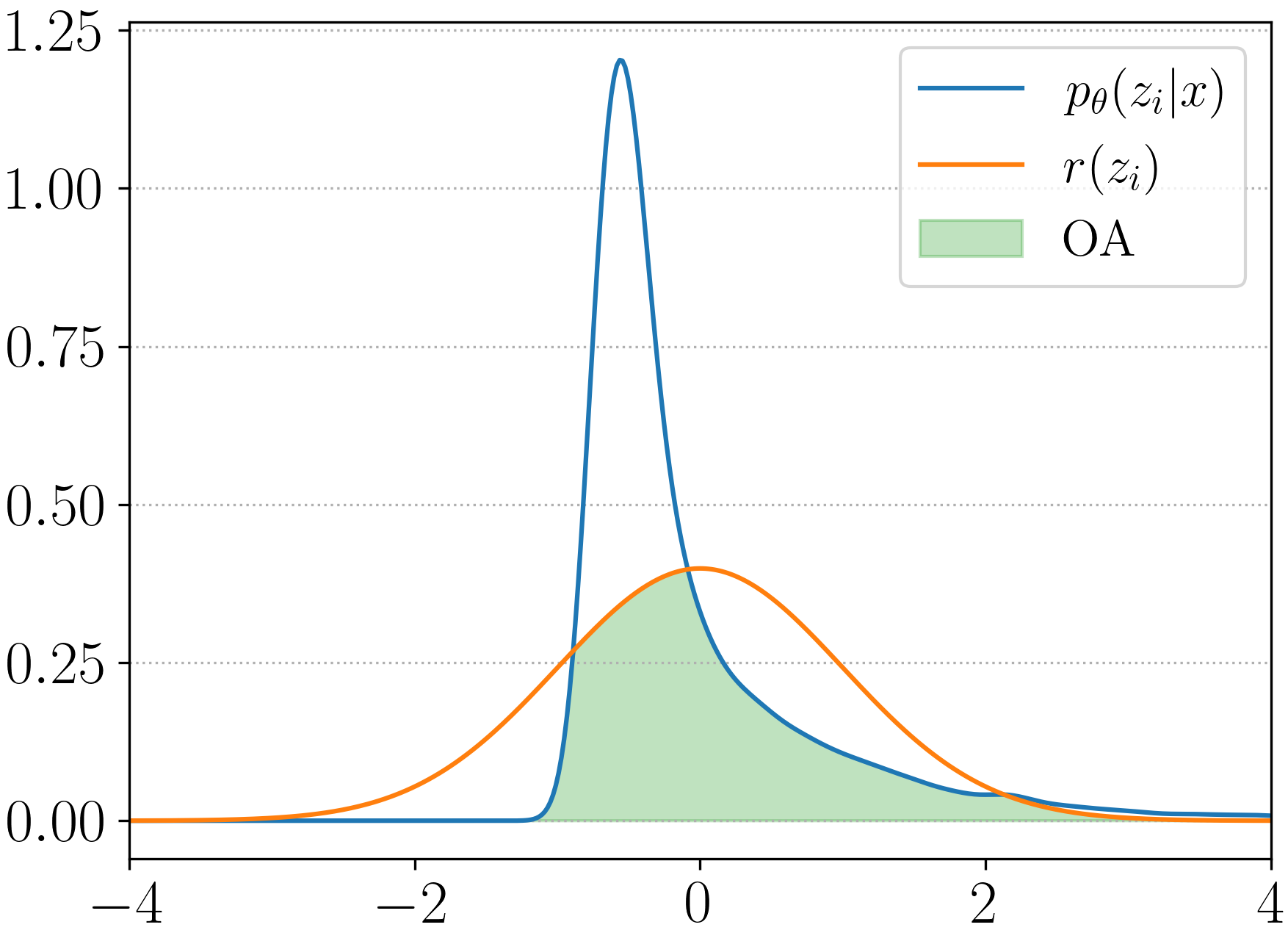}
        }\hspace{1cm}
        \subfloat[P\&L \label{fig:contour_plots_gamma=1.0_PL_OA}]{
		\includegraphics[width=0.22\linewidth]{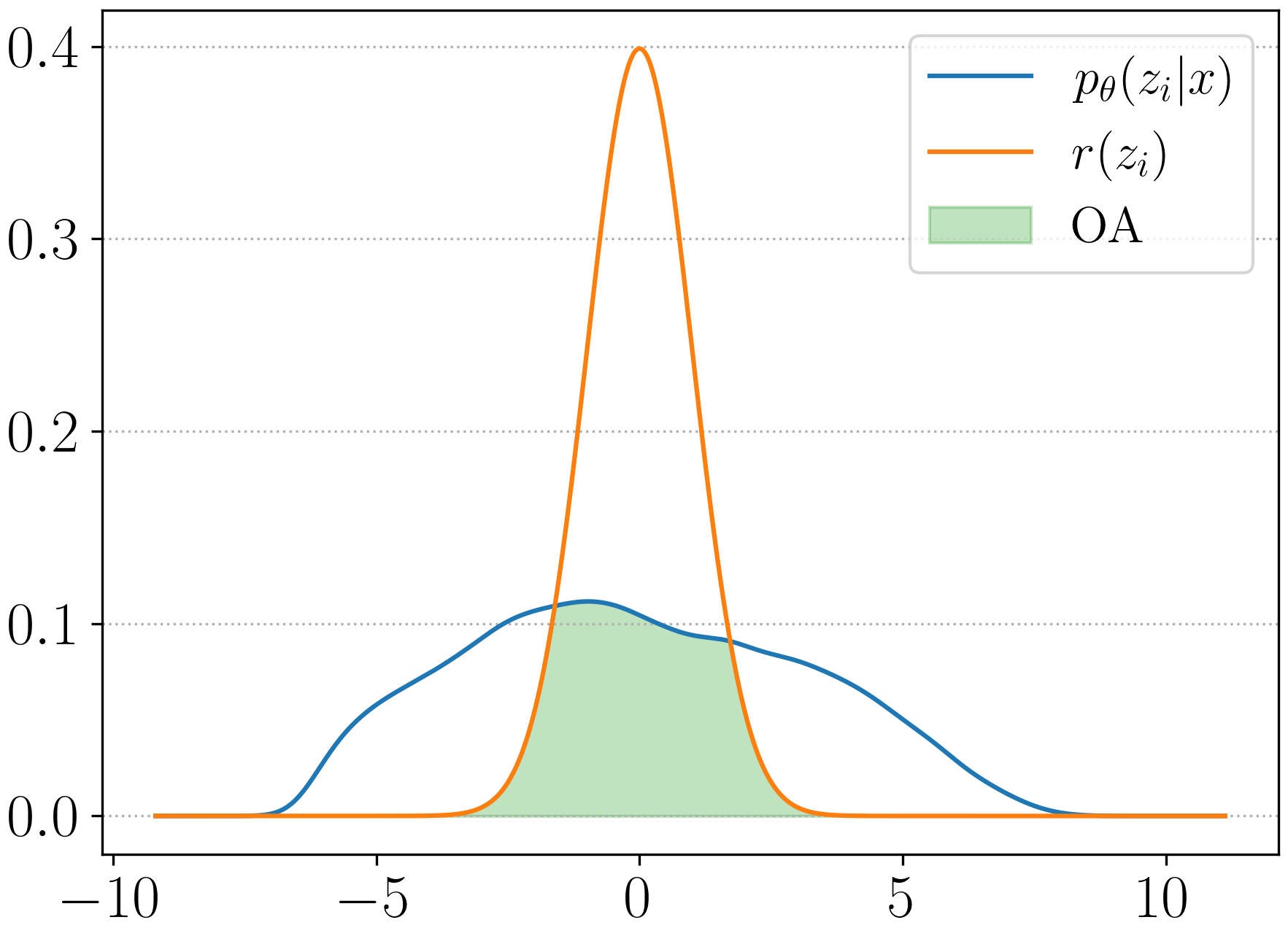}
        }\hspace{1cm}
        \subfloat[PT \label{fig:contour_plots_gamma=1.0_PT_OA}]{
		\includegraphics[width=0.22\linewidth]{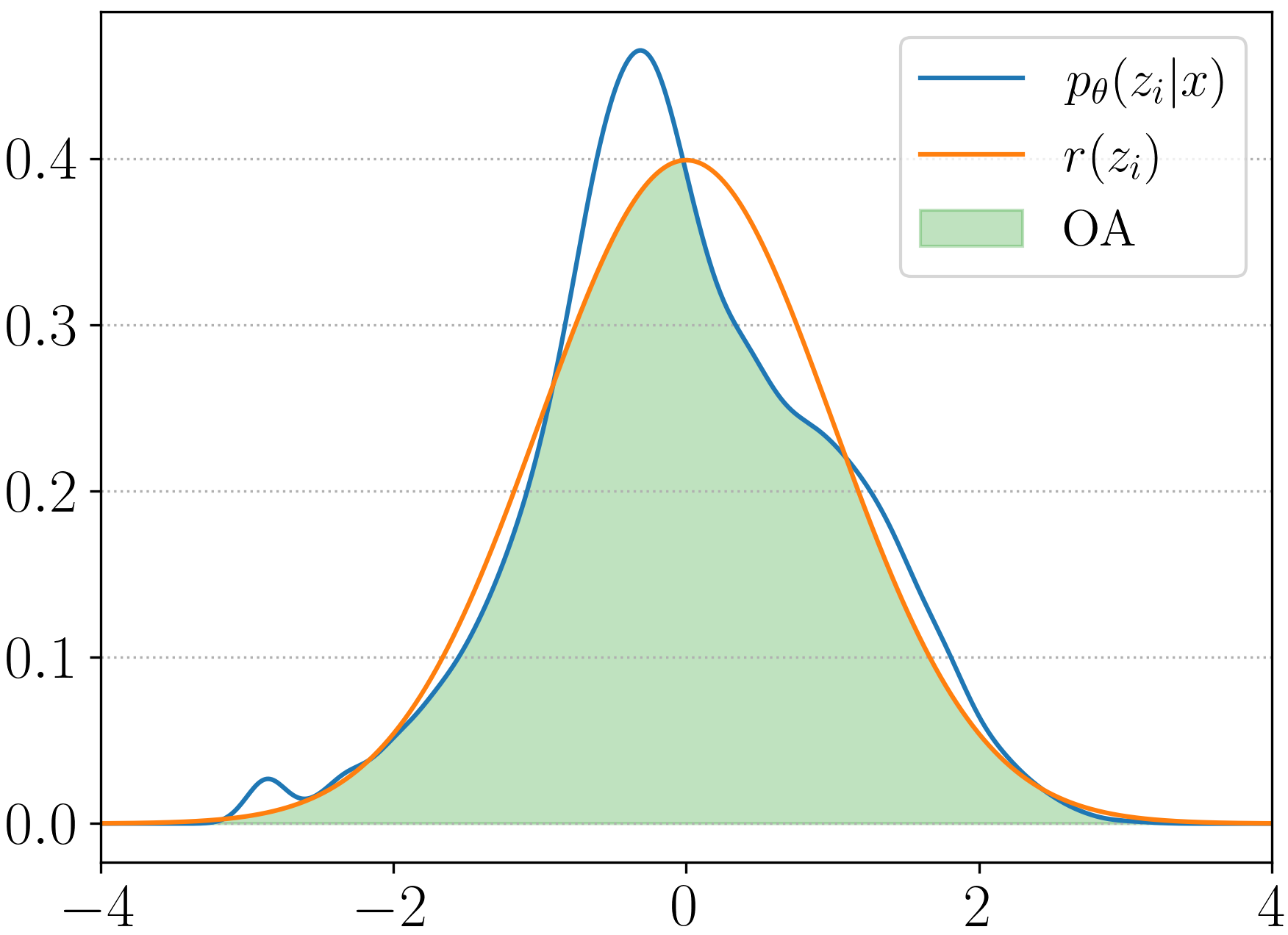}
        }\\
    \caption{\textbf{Contour} attribute regularization with $\beta\rightarrow10^{-3}$ and $\gamma=1$}
\label{fig:contour_plots_gamma=1.0}
\vspace{-0.77em}
\end{figure*}
\begin{table*}[t]
\caption{Spearman's Rank Correlation Coefficients and Regularization Metrics. $(\uparrow)$ the higher the better; $(\downarrow)$ the lower the better.}
\label{tab:results}
\centering
\resizebox{\textwidth}{!}{
    \begin{tabular}{|c|c|cccc|cccc|cccc|}
    \hline
          & \multirow{2}{*}{$\gamma$} & \multicolumn{4}{c|}{\textbf{Contour}}  & \multicolumn{4}{c|}{\textbf{Rhythm Complexity}}  & \multicolumn{4}{c|}{\textbf{Pitch Range}}  \\
    \cline{3-14}
        & & $\rho_\mathrm{s}$ ($\uparrow$) & OA ($\uparrow$) & JSD ($\downarrow$) & MMD ($\downarrow$) & $\rho_\mathrm{s}$ ($\uparrow$) & OA ($\uparrow$) & JSD ($\downarrow$) & MMD ($\downarrow$) & $\rho_\mathrm{s}$ ($\uparrow$) & OA ($\uparrow$) & JSD ($\downarrow$) & MMD ($\downarrow$) \\
    \hline
         NM & \multirow{3}{*}{$10^{-3}$} & 0.4048 & 0.8803 & 0.012652 & 0.3989 & 0.4547 & 0.9055 & 0.007747 & 0.2752 & 0.4494 & 0.8797 & 0.012855 & 0.423175 \\
         P\&L & & 0.5345 & \textbf{0.9888} & \textbf{0.000366} & \textbf{0.0217} & 0.6408 & \textbf{0.9537} & \textbf{0.002618} & \textbf{0.1131} & 0.6646 &  0.8591 & 0.017486 & 0.973104\\
         PT & & 0.5394 & 0.9247 & 0.005627 & 0.2273 & 0.4982 & 0.9133 & 0.007481 & 0.3038 & 0.4028 & \textbf{0.8962} & \textbf{0.009497} & \textbf{0.350608}\\
    \hline
         NM & \multirow{3}{*}{$1.0$} & 0.9994 & 0.4382 & 0.237599 & 256.854 & \textbf{0.9981} & 0.7946 & 0.035963 & 1.9022 & \textbf{0.9982} & 0.7829 & 0.040768 & 1.2870 \\
         P\&L & & \textbf{0.9999} & 0.6041 & 0.121547 & 6.4854 & \textbf{0.9981} & 0.6699 & 0.088264 & 30.3475 & \textbf{0.9982} & 0.7871 & 0.044671 & 14.4203\\
         PT & & 0.9998 & \textbf{0.9224} & \textbf{0.004662} & \textbf{0.0085} & \textbf{0.9981} & \textbf{0.8793} & \textbf{0.013633} & \textbf{0.0523} & \textbf{0.9982} & \textbf{0.9474} & \textbf{0.002732} & \textbf{0.0249} \\
    \hline
    \end{tabular}
}%
\vspace{-0.5em}
\end{table*}


\subsection{Dataset}
\label{ssec:dataset}

We train attribute-controlled models to learn pitch sequence representations of four-bar monophonic melodies. We create a large-scale dataset of monophonic melodies using $176,\!581$ MIDI files from the Lakh MIDI Dataset~\cite{raffelLearningBasedMethodsComparing2016a}.
First, we check whether the original piece contains changes in its time signature. If so, we split the file where these variations occur and keep only the parts with a $4/4$ time signature. Next, we quantize MIDI events to the nearest sixteenth note.
We then define a melody as a sequence of notes within the pitch range of a canonical $88$-key piano and performed by an instrument associated to a valid MIDI program. Considering a melody ended if one bar of silence is found, we extract only melodies that are at least four bars long and contain at least three different pitches. If more than one note is played simultaneously, we follow~\cite{robertsHierarchicalLatentVector2018} and obtain a monophonic sequence by keeping only the note with the highest pitch. Finally, we extract four-bar melodies with a stride of one bar. 
For each melody thus obtained, we compute $13$ musical attributes, including those discussed in Section~\ref{ssec:attributes}.
Melodies are encoded as sequences of $64$ integers in the range $[0, 129]$, with each step corresponding to the MIDI note number if an onset is present or one of two additional tokens for \textit{note off} ($128$) and \textit{note hold} ($129$) events. The corpus is then split into training, validation, and test sets, with the training data augmented through transposition by a random number of semitones in the range of $\pm 1$ octaves. 
The resulting dataset contains about $10$ million unique melodies.\footnote{M.~Pettenò, Aug.~2024, ``4 Bars Monophonic Melodies Dataset (Pitch Sequence),'' Zenodo, doi: \url{https://doi.org/10.5281/zenodo.13369389}}

\subsection{Musical Attributes}
\label{ssec:attributes}

In this work, we focus on three of the $13$ musical attributes included in our dataset: (i)
\textbf{Contour}, which captures the extent of melodic movement by averaging the pitch differences between consecutive notes; 
(ii) \textbf{Rhythm Complexity}, evaluated using Toussaint’s metrical complexity measure~\cite{toussaintMathematicalAnalysisAfrican2002};
(iii) \textbf{Pitch Range}, determined by the difference between the highest and lowest MIDI pitch values in the sequence.

\subsection{Evaluation Metrics}
\label{ssec:metrics}

We evaluate each attribute-controlled model on two aspects: the degree of regularization of the latent dimension $z_i$ onto which the attribute $a$ is encoded and the ability of controlling the output attributes at inference time.
 
On the one hand, a well-regularized latent dimension means that the univariate posterior $p_\theta(z_i|x)$ is as close to $r(z_i)$ as possible.
In practice, we apply kernel density estimation with normal kernels to the empirical posterior, and compare the resulting PDF to the target prior $\mathcal{N}(0,1)$. We measure the (dis)similarity between these two probability distributions with three metrics: (i)
\textbf{Maximum Mean Discrepancy} (\textbf{MMD}) with a polynomial kernel \cite{mittal2021symbolicdiffusion}; (ii) \textbf{Overlapping Area} (\textbf{OA}) as previously done in \cite{lerch2020on, choi2020encoding} to compare musical attribute distributions; and the (iii) \textbf{Jensen–Shannon Divergence} (\textbf{JSD}), a symmetrized and bounded KLD-based similarity measure. 

On the other hand, controllability is assessed through the \textbf{Spearman's rank correlation coefficient} ($\rho_\mathrm{s}$) between the regularized dimension $z_i$ and the attributes $a^\star$ of all $22,\!000$ decoded test sequences. A coefficient $\rho_\mathrm{s}$ close to one indicates that there exists a monotonic relationship between $z_i$ and $a^\star$, which, in turn, means that we can generate samples with a desired attribute by navigating the latent space along the $i$-th dimension.

\subsection{Implementation Details}
\label{ssec:implementation}

We implement the VIB models so as to reconstruct the input melodies.
The encoder and decoder architectures follow the design proposed in \cite{robertsHierarchicalLatentVector2018}.
All models are trained for 40,000 iterations with a batch size of 512. The objective in \eqref{eq:ar_vib_objective}, with cross-entropy as the reconstruction loss, is minimized using Adam. 
The learning rate is decreased exponentially from $10^{-3}$ to $10^{-5}$ with a rate of $0.9999$. 
The hyperparameter $\beta$ is annealed exponentially from $0$ to $10^{-3}$, which encourages the model to focus on reconstructing the sequence with high accuracy in the first part of the training~\cite{robertsHierarchicalLatentVector2018}. 
Additionally, we apply teacher forcing within the bottom-level decoder with a probability following a logistic schedule~\cite{robertsHierarchicalLatentVector2018}.
$\lambda_1$ and $\lambda_2$ are determined for each attribute as described in Section~\ref{sec:methodology}.

To study the effect of different design choices in balancing $\beta$ and $\gamma$, respectively KLD and $\mathcal{L}_\text{AR}$, we train all models with $\gamma=10^{-3}$, i.e., equal to the maximum value of $\beta$, and $\gamma=1$, as in \cite{patiAttributebasedRegularizationLatent2021}. These values are kept fixed throughout the training, as our experiments showed that annealing $\gamma$ in the same way as $\beta$ yields equivalent results.

\section{Results}
\label{sec:results}

In this section, we compare three regularization approaches: the method discussed in Section~\ref{sec:methodology}, denoted here as ``PT'', and those from~\cite{mezzaLatentRhythmComplexity2023} and~\cite{patiAttributebasedRegularizationLatent2021}. Specifically, with ``NM'' we refer to the method in~\cite{mezzaLatentRhythmComplexity2023}, given by~\eqref{eq:beta_vae_default_attr_reg_term}, and with ``P\&L'' we refer to the method in \cite{patiAttributebasedRegularizationLatent2021}, given by~\eqref{eq:beta_vae_sign_attr_reg_term} with $\delta = 10$~\cite{patiAttributebasedRegularizationLatent2021}.

Overall, we train $18$ models, considering three regularization losses (NM, P\&L, PT), three musical attributes (Contour, Rhythm Complexity, Pitch Range), and two different values for the weight $\gamma$ ($1$ and $10^{-3}$).
Table~\ref{tab:results} reports the evaluation metrics discussed in Section~\ref{ssec:metrics} for all such models. 

When $\gamma=10^{-3}$, i.e., when the weighting of the AR loss is comparable to that of the KLD, OA approaches the theoretical maximum of 1, while JSD and MMD are low. This indicates that the minimization of the KLD successfully regularized all latent dimensions, including $z_i$. Conversely, $\rho_\mathrm{s}$ is low, consistently between $0.4$ and $0.66$, i.e., far from the maximum possible value of $1$; this suggests some degree of attribute regularization, yet not enough to ensure fine-grained control over the output attributes in practical applications. 

These phenomena are illustrated for Contour\footnote{Here, we focus on the Contour attribute. Additional results, along with audio examples, are available at \url{https://mpetteno.github.io/box-cox-latent-reg}}
in Fig.~\ref{fig:contour_plots_gamma=0.001}.
For all $22,\!000$ sequences in the held-out test set, Figs.~\ref{fig:contour_plots_gamma=0.001_NM}, \ref{fig:contour_plots_gamma=0.001_PL}, \ref{fig:contour_plots_gamma=0.001_PT} depict $z_i$ against another latent dimension $z_\ell$ that we choose to be the least correlated with $z_i$, with brighter colors corresponding to higher $a^\star$ values and vice versa.
The circular-like distributions of points centered at the origin indicate that, at least approximately, $z_i$ and $z_\ell$ are jointly Gaussian, as prescribed by the KLD with a normal prior. 
The lack of a smooth color gradient along $z_i$, however, indicates poor AR, corroborating the values of $\rho_\mathrm{s}$ ranging from $0.40$ to just below $0.54$ in Table~\ref{tab:results}.
Figs.~\ref{fig:contour_plots_gamma=0.001_NM_OA}, \ref{fig:contour_plots_gamma=0.001_PL_OA}, \ref{fig:contour_plots_gamma=0.001_PT_OA} show the comparison between $p_\theta(z_i|x)$ (in blue) and $r(z_i)$ (in orange), as well as the overlap between the two distributions (in green). These plots provide visual evidence of the similarity measured by OA, JSD, and MMD. Fig.~\ref{fig:contour_plots_gamma=0.001_PL_OA} also reveals that, with $\gamma=10^{-3}$, P\&L exhibits a better synergy with the KLD compared to NM and PT, possibly because \eqref{eq:beta_vae_sign_attr_reg_term} enforces a monotonic rather than pointwise error constraint. 

When $\gamma=1$, instead, AR appears to govern the training dynamics. As such, $\rho_\mathrm{s}$ turns out consistently over $0.99$, indicating an almost perfect monotonic relationship.
On the downside, the regularization metrics for NM and P\&L significantly worsen. This can be observed for Contour in Figs.~\ref{fig:contour_plots_gamma=1.0_NM} and \ref{fig:contour_plots_gamma=1.0_PL}, where despite a distinct gradient from dark to bright colors, the univariate latent distributions appear to be far from jointly Gaussian. Likewise, Fig.~\ref{fig:contour_plots_gamma=1.0_NM_OA} shows that NM encodes an exponential distribution onto $z_i$, thus drastically reducing the OA, while Fig.~\ref{fig:contour_plots_gamma=1.0_PL_OA} reveals that P\&L leads to monotonic yet sub-Gaussian attribute encoding.

This means that, when using these regularization losses, it is not sufficient to sample the prior $r(z)$ at inference time, but one should either have prior knowledge of the true attribute distribution or empirically estimate the latent marginal from representative encodings of real-world data before being able to apply meaningful manipulation.
In contrast, when it comes to the PT regularization method, OA, JSD, and MMD are comparable if not better with $\gamma=1$ than with $\gamma=10^{-3}$, all while maintaining a Spearman's rank correlation coefficient over $0.99$. Therefore, PT-regularized models appear to offer greater flexibility when it comes to hyperparameter tuning, allowing one to prioritize controllability without penalizing regularization, and thus enabling inference by sampling the prior and navigating the latent space as in a regular VIB.


\section{Conclusions}
\label{sec:conclusions}

In this work, we explored the challenges of jointly minimizing regularization loss functions in variational information bottleneck models for symbolic music generation. We examined the trade-off between the Kullback-Leibler divergence and existing attribute-regularization losses, highlighting how an imbalance between these objectives can compromise either controllability or adherence to the prior along the target dimensions. Our findings indicate that existing approaches struggle to balance these constraints, resulting in either poor control over musical attributes or deviations from the desired latent structure. However, experimental results demonstrate that invertible mappings between attributes and latents based on the Box-Cox power transform can alleviate this issue, enabling both improved controllability and robust latent regularization with minimal hyperparameter tuning. Future work includes learning transformation parameters via backpropagation, extending the method to handle multiple attributes, and validating the framework across different signal domains.





\bibliographystyle{IEEEtran}
\bibliography{biblio}

\end{document}